\documentclass[twoside]{article}

\usepackage[accepted]{aistats2019}
\usepackage[greek, english]{babel}
\usepackage[utf8]{inputenc} %
\usepackage{booktabs}       %
\usepackage{amsfonts}       %

\usepackage{csquotes} %

\usepackage[dvipsnames,usenames]{xcolor}
\definecolor{NavyBlue}{cmyk}{0.94,0.54,0,0}
\definecolor{DarkBlue}{rgb}{0,0.2,0.6}

\usepackage[colorlinks=true, linkcolor=RawSienna,citecolor=DarkBlue,urlcolor=DarkBlue,bookmarks=false]{hyperref}

\usepackage{xspace}

\usepackage[
    backend=biber,
    style=authoryear-comp,
    natbib=true,    %
    url=false, 
    dashed=false,	%
	isbn=false,
    doi=false,
	maxcitenames=2, %
	maxbibnames=8, %
	giveninits=true, %
	uniquename=false,
	uniquelist=false,
    eprint=true
]{biblatex}
\citetrackerfalse 
\usepackage{bbm}          %
\usepackage{bm}          %

\usepackage{amssymb,amsmath,amsthm, mathtools}
\usepackage{amscd}
\usepackage{stmaryrd}

\usepackage{caption, subcaption}%
\usepackage[export]{adjustbox}

\newcommand{\sref}[1]{\S\ref{#1}}

\usepackage{siunitx} %

\usepackage{algorithm}
\usepackage{algorithmic}

\usepackage[colorinlistoftodos]{todonotes}
\newtheorem{theorem}{Theorem}[section]

\newtheorem{lemma}[theorem]{Lemma}
\newtheorem{lemma*}{Lemma}

\makeatletter
\renewcommand*\env@matrix[1][*\c@MaxMatrixCols c]{%
  \hskip -\arraycolsep
  \let\@ifnextchar\new@ifnextchar
  \array{#1}}
\makeatother

\def\R{\mathbb{R}}
\def\K{\mathbb{K}}
\def\N{\mathbb{N}}
\def\Q{\mathbb{Q}}

\def\A{\mathbf{A}}
\def\B{\mathbf{B}}
\def\C{\mathbf{C}}
\def\D{\mathbf{D}}
\def\E{\mathbf{E}}
\def\F{\mathbf{F}}
\def\G{\mathbf{G}}
\def\H{\mathbf{H}}
\def\I{\mathbf{I}}
\def\K{\mathbf{K}}
\def\L{\mathbf{L}}
\def\M{\mathbf{M}}
\def\P{\mathbf{P}}
\def\U{\mathbf{U}}
\def\V{\mathbf{V}}
\def\Y{\mathbf{Y}}
\def\X{\mathbf{X}}
\def\W{\mathbf{W}}
\def\w{\mathbf{w}}
\def\x{\mathbf{x}}
\def\v{\mathbf{v}}
\def\y{\mathbf{y}}
\def\z{\mathbf{z}}
\def\c{\mathbf{c}}
\def\e{\mathbf{e}}
\def\f{\mathbf{f}}
\def\g{\mathbf{g}}
\def\p{\mathbf{p}}
\def\q{\mathbf{q}}
\def\r{\mathbf{r}}
\def\b{\mathbf{b}}
\def\s{\mathbf{s}}
\def\a{\mathbf{a}}
\def\d{\mathbf{d}}
\def\u{\mathbf{u}}
\def\0{\mathbf{0}}
\def\ones{\mathbbm{1}}
\def\cX{\mathcal{X}}
\def\cY{\mathcal{Y}}

\def\cF{\mathcal{F}}

\def\st{\medspace|\medspace}

\DeclarePairedDelimiterX{\inner}[2]{\langle}{\rangle}{#1, #2}

\DeclareDocumentCommand{\tr}{s m}{%
  \operatorname{tr}%
  \IfBooleanTF{#1}%
    {#2}%
    {\left(#2\right)}%
}%
\DeclareDocumentCommand{\diag}{s m}{%
  \operatorname{diag}%
  \IfBooleanTF{#1}%
    {#2}%
    {\left(#2\right)}%
}%

\def\cov{\text{cov}}

\DeclareMathOperator*{\argmin}{argmin}
\DeclareMathOperator*{\argmax}{argmax}

\DeclarePairedDelimiterX{\infdivx}[2]{(}{)}{%
  #1\;\delimsize\|\;#2%
}

\newcommand{\suchthat}{\;\ifnum\currentgrouptype=16 \middle\fi|\;}

\newcommand{\ra}[1]{\renewcommand{\arraystretch}{#1}}

\newenvironment{itemize*}%
  {\begin{itemize}%
  \vspace{-0.5cm}
    \setlength{\itemsep}{0pt}%
    \setlength{\parskip}{0pt}}%
  {\end{itemize}}
\newenvironment{enumerate*}%
{\begin{enumerate}
    \vspace{-0.5cm}
    \setlength{\itemsep}{0pt}%
    \setlength{\parskip}{0pt}}%
  {\end{enumerate}}

\newcommand{\procru}{\textsc{Procrustes}\xspace}

\newcommand{\En}{\textsc{En}\xspace}
\newcommand{\Es}{\textsc{Es}\xspace}
\newcommand{\De}{\textsc{De}\xspace}

\newcommand{\Fr}{\textsc{Fr}\xspace}
\newcommand{\It}{\textsc{It}\xspace}
\newcommand{\Ru}{\textsc{Ru}\xspace}

  {\list{}{\leftmargin=#1\rightmargin=#1}\item[]}%
  {\endlist}

 \newtheorem{innercustomlemma}{Lemma}
 \newenvironment{customlemma}[1]{\renewcommand\theinnercustomlemma{#1}\innercustomlemma}{\endinnercustomlemma}

\begin{document}

\twocolumn[

\aistatstitle{Towards Optimal Transport with Global Invariances}

\aistatsauthor{ David Alvarez-Melis \And Stefanie Jegelka \And  Tommi S. Jaakkola  }
\aistatsaddress{ CSAIL, MIT \And CSAIL, MIT \And CSAIL, MIT } 

]

\begin{abstract}

    Many problems in machine learning involve calculating correspondences between sets of objects, such as point clouds or images. Discrete optimal transport provides a natural and successful approach to such tasks whenever the two sets of objects can be represented in the same space, or at least distances between them can be directly evaluated. Unfortunately neither requirement is likely to hold when object representations are learned from data. Indeed, automatically derived representations such as word embeddings are typically fixed only up to some global transformations, for example, reflection or rotation. As a result, pairwise distances across two such instances are ill-defined without specifying their relative transformation. In this work, we propose a general framework for optimal transport in the presence of latent global transformations. We cast the problem as a joint optimization over transport couplings and transformations chosen from a flexible class of invariances, propose algorithms to solve it, and show promising results in various tasks, including a popular unsupervised word translation benchmark.

\end{abstract}

\section{Introduction}

Optimal transport (OT) plays dual roles across machine learning applications. First, it provides a well-founded, geometrically driven approach to realizing correspondences between sets of objects such as shapes in different images. Such correspondences can be used for image registration \citep{haker2001optimal} or to interpolate between them \citep{Solomon2015Convolutional}. More generally, OT extends to problems such as domain adaptation where we wish to transport a set of labeled source points to the realm of the target task \citep{Courty2017Optimal, Courty2017Joint}. Second, the transportation problem induces a theoretically well-characterized distance between distributions. This distance is expressed in the form of a transport cost and serves as a natural population difference measure, which can be exploited as a source of feedback in adversarial training \citep{Arjovsky2017Wasserstein, Bousquet2017}. Our focus in this paper is on the optimal coupling mediating the transport, i.e., realizing the latent correspondences between the objects.

A key limitation of classic OT is that it implicitly assumes that the two sets of objects in question are represented in the same space, or at least that meaningful pairwise distances between them can be computed. This is not always the case, especially when the objects are represented by learned feature vectors. For example, word embedding algorithms operate at the level of inner products or distances between word vectors, so the representations they produce can be arbitrarily rotated, sometimes even for different runs \emph{of the same algorithm on the same data}. Such global degrees of freedom in the vector representations render direct pairwise distances between objects across the sets meaningless. Indeed, OT is \emph{locally greedy} as it focuses on minimizing individual movement of mass, oblivious to global transformations. As a concrete example, consider two identical sets of points where one set is subjected to a global rotation. The optimal transport coupling evaluated between the resulting sets may no longer recover the correct correspondences (Figure~\ref{fig:synth_results}).

When the global transformation is known or can be easily estimated, it can be incorporated in the computation of pairwise distances, thereby enabling the use of traditional OT. Unfortunately, only the \emph{type} of underlying transformation (e.g., rotation) is typically known, not the actual realization. In such cases, we would like the optimal transport problem to also find the best latent transformation along with the optimal coupling. In other words, we seek a formulation of OT that remains invariant under global transformations.

In this work, we propose a generalization of discrete optimal transport that incorporates global invariances directly into the optimization problem. While the driving motivation is invariance to rigid transformations (arguably the most common case in practice), our framework is very general and allows various other types of invariances to be encoded. Moreover, our approach unifies previous methods for fusing OT with global transformations such as Procrustes mappings \citep{Rangarajan1997Softassign, zhang2017earth, grave2018unsupervised}, and reveals unexpected connections to the Gromov-Wasserstein distance \citep{memoli2011gromov}.

The main contributions of this work are thus:
\begin{itemize}
	\vspace{-.3cm}
	\itemsep-0.3em
	\item A novel formulation of the discrete optimal transport problem that allows for global geometric invariances to be incorporated into the objective;
	\item Design and analysis of efficient algorithms for this general class of problems;
	\item An application of the framework to the problem of unsupervised word translation, yielding performance comparable to state-of-the-art methods at a fraction of their computational cost.
\end{itemize}

\section{Related Work}\label{sec:related}

The general problem of unsupervised estimation of correspondence between sets of features is well-studied and arises in various fields under different names, such as manifold alignment \citep{wang2009manifold}, feature set matching \citep{grauman2005pyramid} and feature correspondence finding \citep{torresani2008feature}. Here, we focus the discussion on related methods that combine soft correspondences (such as those produced by OT) with explicit space alignment.  

Perhaps the earliest such approach is by \citet{Rangarajan1997Softassign}, who derive a framework to establish correspondences between shapes that rejects non-homologies (e.g., rotations) based on an entropy-regularized version of the OT problem. The resulting \emph{Softassign Procrustes Algorithm} proceeds iteratively by alternating between estimating optimal rotations and performing Sinkhorn iterations. This approach, however, only considers rotations, and is tailored to 2D data, where rotations can be easily parametrized.

More recently, \citet{zhang2017earth} propose combining OT with Procrustes alignment to find correspondences between word embedding spaces. They initialize their orthogonal mapping using an adversarial training phase, much like \citet{conneau2018word}, and solve the optimization problem with alternating minimization. Our approach, on the other hand, does not rely on neural network initialization, instead leveraging a convexity annealing scheme that leads to smooth convergence, with little sensitivity to initialization. 

Concurrently with our work, \citet{grave2018unsupervised} tackle the word embedding alignment task with an approach similar to that of \citet{zhang2017earth}, combining Wasserstein distances (an instance of OT) and Procrustes alignment. Their approach differs from \citet{zhang2017earth} in how they scale up optimization, by relying on a stochastic Sinkhorn solver \citep{genevay2018learning}, and in how they initialize it, by solving a convex relaxation of the original problem.\footnote{Interestingly, this relaxation corresponds to a hybrid of two instances of our framework: optimizing a Frobenius-norm objective (\sref{sub:p2}) over orthogonal matrices (\sref{sub:pinfty}).}

Although driven by a similar motivation (word embedding alignment) and relying on similar principles (joint optimization of transport coupling and feature mapping) as the work of \citet{zhang2017earth} and \citet{grave2018unsupervised}, our approach differs from them in several aspects. First, it allows for more general types of invariance classes (characterized as  Schatten $\ell_p$-norm balls), subsuming orthogonal invariance considered in prior work as a special case. Second, we dispense with the need for any ad-hoc initialization by introducing instead a convexity-annealing approach to optimization. Third, our approach remains robust to the choice of entropy regularization parameter $\lambda$. 

Compared to methods that directly solve a Procrustes problem from a few known correspondences \citep{zhang2016ten, artetxe2017learning} or by generating pseudo-matches through an initial unsupervised step \citep{conneau2018word, Artetxe2018Robust}, OT allows for more flexible correspondence estimation.

A different generalization of the OT problem aimed at overcoming lack of intrinsic correspondence between spaces is the Gromov-Wasserstein (GW) distance \citep{memoli2011gromov}. It has been recently applied to various correspondence problems, including shape interpolation \citep{solomon2016entropic, peyre2016gromov} and unsupervised word translation \citep{alvarez-melis2018gromov}. While our framework recovers this distance in certain scenarios (see \sref{sub:p2}), it is best understood as a compromise between the classic formulation of OT that requires the spaces to be fully registered, and the GW distance, which completely forgoes explicit computation of distances across spaces, relying instead on comparison of intra-space similarities. Thus, our approach is best suited to tasks where distances across spaces can be computed, but are meaningful only if made invariant to some latent transformation. A further difference with the usual OT and GW distances is that our approach produces, as intrinsic part of optimization, a global mapping which can be used to map \textit{out-of-sample points} across spaces.

%

\section{Feature Space Correspondences}

\paragraph{Notation.} We denote vectors and matrices with bold font (e.g., $\x$, $\X$), their entries without it ($x_i, X_{ij}$), and sets as $X$, $Y$. We use super-indices to enumerate vectors, and subindices to denote their entries. For matrices $\A, \B$, let $\inner{\A}{\B} = \sum_{i,j} [\A]_{ij}[\B]_{ij}$ denote their Frobenius inner product, and $\| \cdot\|_{*}$ the nuclear (trace) norm. We denote by $\mathcal{O}_n$ the set of orthogonal matrices of order $n$, and finally, let $\llbracket n \rrbracket \triangleq \{1,\dots,n\}$.

\subsection{Problem Setting}
Suppose we are given two sets of examples, $X=\{\x^{(i)}\}_{i=1}^n$ and $Y=\{\y^{(j)}\}_{j=1}^m$, drawn from potentially distinct feature spaces $\mathcal{X} \subset \R^{d_x}$ and $\mathcal{Y} \subset\R^{d_y}$. Our goal is to learn correspondences between $X$ and $Y$, in the challenging case where no prior instance-wise correspondences are known---i.e., the problem is fully unsupervised---\textit{and} the spaces $\cX$ and $\cY$ are unregistered---i.e., the \textit{global} correspondence between them is unknown too. 

This task can be thought of as consisting of two sub-problems: (i)  finding a global alignment of spaces $\cX$ and $\cY$, e.g., via a mapping $T:\mathcal{X} \rightarrow \mathcal{Y}$ such that $\| T(\x) - \y \|$ is small for every correspondence pair $(\x,\y)$, and (ii) finding correspondences between the items in $X$ and $Y$, via an assignment $\mathcal{A}: \llbracket n \rrbracket\mapsto \llbracket m \rrbracket$ such that $\x^{(i)} \rightarrow \y^{(j)}$ if these points are in correspondence. Individually, these two problems are well-studied and understood. Below, we briefly discuss popular approaches to solve them. In Section~\ref{sec:approach}, we show how to combine them to enforce invariances in the optimal transport problem, leading to a flexible class of problems that can be solved efficiently.

\subsection{Space alignment from paired samples} %
\label{sub:subsection_name}
Assume for now\footnote{We discuss how to relax these in Appendix.} that $m=n$ and $d_x = d_y$. Given paired samples from the two domains, consider matrices $\X\in\R^{d\times n}$ and $\Y\in\R^{d\times n}$, whose columns correspond to these paired elements. The problem of finding the best mapping $T$ that maps the target samples to the source ones can be cast as
\begin{equation*}
	\min_{T \in \mathcal{F}} \| \X - T(\Y) \|^2
\end{equation*}
where $\mathcal{F}$ is some class of functions and $\| \cdot\|$ is typically taken to be the Frobenius norm $\|\A\|_F = \sqrt{\sum_{i,j}|a_{ij}|^2}$.
Naturally, the choice of space $\cF$ will determine the difficulty of finding $T$ as well as the quality of the alignment implied by it.

The classic \emph{Orthogonal Procrustes problem} restricts $\cF$ to be rigid (rotation and reflection) transformations---i.e., orthogonal matrices:
\begin{equation}\label{eq:procrustes}
	\min_{\P \in \mathcal{O}_n} \| \X - \P\Y \|_F^2
\end{equation}

Despite its simplicity, Procrustes analysis is a powerful tool used in various applications, from statistical shape analysis \citep{goodall1991procrustes} to market research and others \citep{gower2004procrustes}. Its main advantage is that it has a closed-form solution in terms of a singular value decomposition (SVD) \citep{schonemann1966generalized}. Namely, given an SVD, say $\U\Sigma\V^\top$, of $\X\Y^\top$, the orthogonal matrix minimizing problem \eqref{eq:procrustes} is $\P^* =  \U\V^\top$, which is a direct consequence of a well-known approximation property of the SVD:
\begin{lemma}\label{lemma:procrustes_general}
	If $\A \in \R^{n\times m}$ and $\A=\U\Sigma\V^\top$ is an SVD of $\A$, then
	 $\argmax_{\P \in \mathcal{O}_n} \inner{\P}{\A} =  \U\V^\top$.
	 \begin{proof}
		 \vspace{-.25cm}
		  This is a particular case of Lemma~\ref{lemma:svd_general}. All other proofs are provided in the Appendix.
	 \end{proof}
\end{lemma}

We emphasize that the Procrustes problem \eqref{eq:procrustes} crucially requires the columns of $\X$ and $\Y$ be paired, making it an intrinsically \emph{supervised} approach.  Thus, its application to the problem of feature alignment requires either an---ideally small---initial set of true paired examples or a method to generate them.

\subsection{Correspondences between aligned spaces} %
\label{sub:optimal_transport}

Optimal transport is an appealing approach for finding correspondences between feature vectors in a fully unsupervised way. Besides strong theoretical foundations and fast algorithms, it is ideal for this setting, as it is fully unsupervised, inferring correspondences by leveraging the geometry of these collections.

The OT problem considers two measures $\mu$ and $\nu$ over spaces $\cX$ and $\cY$, and a transport cost $c: \cX \times \cY \rightarrow \R^+$. It seeks to minimize the cost of \textit{transporting} measure $\mu$ onto $\nu$. In its original discrete formulation, $\mu$ and $\nu$ are taken as empirical distributions:
\begin{equation}
	\mu = \sum_{i=1}^n p_i\delta_{\x^{(i)}}, \quad \nu = \sum_{j=1}^m q_j\delta_{\y^{(j)}}
\end{equation}
so all relevant pairwise costs can be represented as a matrix $\C \in \R^{n \times m}$. Monge's formulation seeks a transport map $T:\cX \rightarrow \cY$ that assigns each point $\x^{(i)}$ to a single point $\y^{(j)}$, and which pushes the mass of $\mu$ toward that of $\nu$ (i.e., $T_\sharp\mu=\nu$), while minimizing:
\begin{equation}\label{monge_ot}
	\min_{T} \left\{ \sum_{i=1}^n c(\x^{(i)},T(\x^{(i)})) \medspace : \medspace T_\sharp\mu=\nu \right\}
\end{equation}
Since the solution to this problem is not guaranteed to exist, a relaxation of this problem by Kantorovich considers instead ``soft'' assignments defined in terms of probabilistic \textit{couplings} $\Gamma \in \R_{+}^{n\times m}$ whose marginals recover $\mu$ and $\nu$. Formally, Kantorovich's formulation seeks $\Gamma$ in the transportation polytope:
\begin{equation}\label{transport_poly}
	\Pi(\p,\q) = \{ \Gamma \in \mathbb{R}^{n \times m}_+ \st \Gamma \mathbf{1} = \p, \medspace \Gamma^\top\mathbf{1} = \q \}
\end{equation}
that solves
\begin{equation}\label{eq:original_OT}
	\min_{\Gamma \in \Pi(\p, \q)} \langle \Gamma, \C \rangle .
\end{equation}

The discrete optimal transport (DOT) problem is a linear program. Practical algorithms to solve it include Orlin's algorithm and interior point methods, both of which have $O(n^3 \log n)$ complexity \citep{pele2009fast}. However, adding an entropy regularization term leads to much more efficient algorithms:

\begin{equation}\label{eq:entropy_reg_ot}
\min_{\Gamma \in \Pi(\p,\q)}\; \langle \Gamma, \C \rangle - \frac{1}{\lambda}H(\Gamma).
\end{equation}
This is a strictly convex optimization problem, whose solution has the form $\Gamma^* = \diag{\u}\K\diag{\v}$, with $\K = e^{-\frac{\C}{\lambda}}$ where the exponential is computed entry-wise \citep{Peyre2018Computational}, and can be obtained efficiently via the Sinkhorn-Knopp algorithm, an iterative matrix-scaling procedure \citep{cuturi2013sinkhorn}. Besides computational motivation, introducing this regularization often leads to better empirical performance in applications where having denser correspondences is beneficial, e.g., in settings where the support points correspond to noisy features \citep{AlvarezMelis2018Structured}.

Unfortunately, naive application of DOT often fails in applications where the two spaces $\cX$ and $\cY$ are not \emph{registered}: i.e., when there is no a priori meaningful notion of distance between them. As stated before, this is likely to be the case when the feature spaces are learned. In such cases, even if the feature spaces are of the same dimensionality, the naive choice of cost function $c(\x,\y) = \|\x - \y \|$ is flawed, as there is no guarantee that these two spaces are coherent, i.e., that their coordinate axes are in correspondence. To remedy this situation, we seek to make optimal transport invariant to global transformations of the space. In the next section, we do so for a general class of invariances.

\section{Transporting with Global Geometric Invariances}\label{sec:approach}

Our goal in this work is to extend the DOT problem \eqref{eq:original_OT} to enforce invariance with respect to certain classes of transformations. Formally, we assume there exists an unknown function $f$ in a pre-specified class $\cF$ which characterizes the global correspondence between spaces $\cX$ and $\cY$, i.e., for which $\cX = \{\x \in \R^d \suchthat \x = f(\y), \medspace \y \in \cY\}$. Naturally, the choice of $\cF$ should be informed by the application domain. 

\begin{figure}
	\centering
	\includegraphics[width=0.95\linewidth]{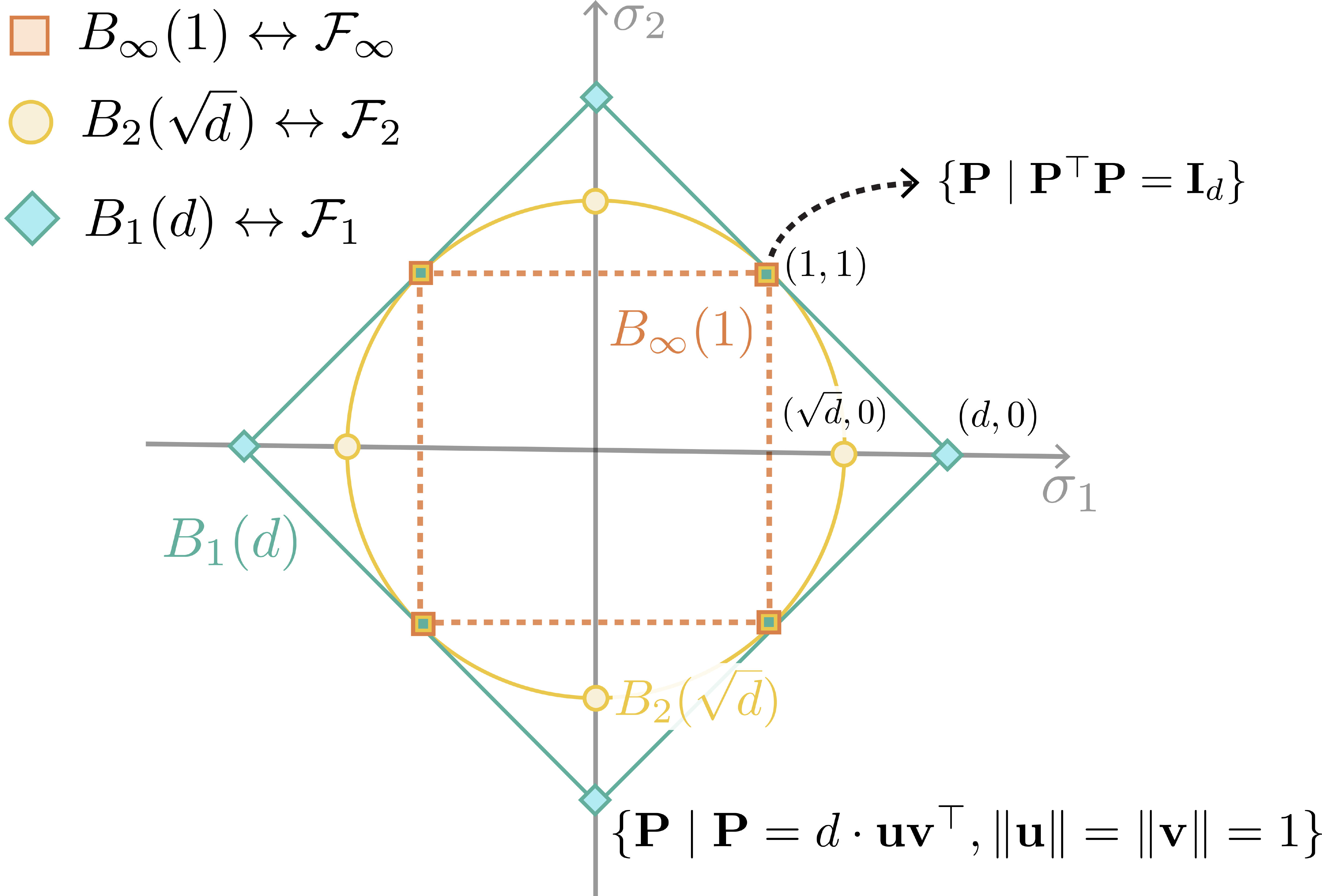}
	\caption{\textbf{Schatten-norm invariance classes}. The depicted $\ell_p$-norm balls in singular value space correspond to matrix invariance classes $\cF_p$. The radius is chosen so as to include the identity matrix ($\bm{\sigma} = [1,1]$). For linear objectives, solutions when optimizing over $\cF_{\infty}$ and $\cF_{1}$ can be found on the extreme points of their respective constraint spaces: orthogonal matrices for the former and rank-one matrices for the latter.}\label{fig:schatten}
\end{figure}

In this setting, given collections $\{\x^{(i)}\}_{i=1}^n$ and $\{\y^{(j)}\}_{j=1}^m$ and associated empirical measures $\mu,\nu$, we seek to simultaneously find the best global transformation of the space (within $\cF$) and the best local correspondences between the two collections, as defined by the optimal transport problem. In other words, we seek to jointly optimize $f \in \mathcal{F}$ and $\Gamma \in \Pi(\p,\q)$ to minimize the transportation cost between the two collections. Formally, given an invariance set $\mathcal{F}$, for $f\in \mathcal{F}$ let $f(\Y)$ denote the matrix of size $d \times m$ whose columns are $f(\y^{(j)})$. The problem we wish to solve is
\begin{equation}\label{eq:original_formulation}
		\min_{\Gamma \in \Pi(\p,\q)} \min_{f \in \cF} \inner{\Gamma}{C(\X,f(\Y))}.
\end{equation}
where $[C(\X,f(\Y))]_{ij}=c(\x^{(i)}, f(\y^{(j)}))$. In this work, we focus on invariances defined by linear operators with bounded norm:
\begin{equation}
	\mathcal{F}_p \triangleq \{ \P\in\R^{d\times d} \suchthat \|\P\|_p \leq k_p  \},
\end{equation}
where $\| \cdot \|_p$ is the Schatten $\ell_p$-norm, that is,  $\|\P\|_p = \|\sigma(\P)\|_p$ where $\sigma(\P)$ is a vector containing the singular values of $\P$. In addition, $k_p$ is a norm- and problem-dependent constant.\footnote{In the most common case, $k_p$ would be chosen to ensure the identity mapping is contained in this set.} This choice of invariance sets follows both modeling and computational motivations. As for the former, Schatten norms allow for immediate interpretation of the elements of $\cF_p$ in terms of their spectral properties (Fig.~\ref{fig:schatten}). For example, choosing $p=1$ encourages solutions with sparse spectra (e.g., projections, useful when the support of one of the two distributions is known to be contained in a lower-dimensional subspace), while $p=\infty$ instead seeks solutions with uniform spectra (e.g., unitary matrices, thus enforcing invariance to rigid transformations). Intermediate values of $p$ interpolate between these two extremes. Interestingly, the choice $p=2$ recovers a recent popular generalization of the optimal transport problem motivated by a similar goal: the Gromov-Wasserstein distance \citep{memoli2011gromov}, as we show in Section~\ref{sub:p2}. Thus, the proposed Schatten invariance framework offers significant flexibility. In terms of computation, Schatten norms exhibit various desirable properties, such as unitary invariance, submultiplicativity, and easy characterization via duality, all of which play an important role in deriving efficient optimization algorithms below.

Here, we formulate the problem for the case where the ground metric $c$ is the squared euclidean distance, i.e, $c(\x,\y) = \|\x-\y\|_2^2$, which is arguably the most common choice in practice. With this choice of ground metric, let $\u,\v$ be vectors with entries $u_i = \|\x^{(i)}\|_2^2$, and $v_j = \|\P\y^{(j)}\|_2^2$ respectively. Then, it is easy to show that \eqref{eq:original_formulation} becomes:
\begin{equation}\label{eq:euclidean_formulation}
		\max_{\Gamma \in \Pi(\p,\q)}\max_{\P \in \cF}\;  2\inner{\Gamma}{\X^\top\P\Y} - \inner{\u}{\p} - \inner{\v}{\q}
\end{equation}
This objective has a clear interpretation. The first term, which can be equivalently written as $\inner{\X\Gamma}{\P\Y}$, measures agreement between $\X\Gamma$, the source points mapped according to the barycentric mapping implied by $\Gamma$, and $\P\Y$, the target points mapped according to $\P$. The other two terms, which can be interpreted as empirical expectations $\hat{\mathbb{E}}_{\x \sim \mu}\|\x\|_2^2$ and $\hat{\mathbb{E}}_{\y \sim \nu}\|\P\y\|_2^2$, act as a counterbalance, normalizing the objective and preventing artificial maximization of the similarity term by arbitrary scaling of the mapped vectors. 

In general, Problem \eqref{eq:euclidean_formulation} is not jointly concave on $\P$ and $\Gamma$, but it is concave in either variable if the other one is fixed. Hence, we can solve it via alternating maximization on $\P$ and $\Gamma$. Since only the first term depends on $\Gamma$, solving for this variable for a fixed $\P$ is a usual OT problem, for which we discuss optimization in Section~\ref{sub:optim}. On the other hand, for a fixed $\Gamma$ the problem is a concave maximization over a compact and convex set, which can be solved efficiently with Frank-Wolfe-type algorithms since projecting onto Schatten norm balls is tractable \citep{jaggi2013revisiting}. While this approach provides a tractable way to solve problem \eqref{eq:euclidean_formulation} in general, we show that under conditions that often hold practice, optimization is much simpler. This simplification relies on eliminating the dependence on $\P$ of the third term in problem \eqref{eq:euclidean_formulation}:
\begin{lemma}\label{lemma:simplifying_conditions}
	Under either of the conditions
	\begin{enumerate}
		\itemsep0em
		\vspace{-.25cm}
		\item $\forall \P \in \mathcal{F}$, $\P$ is angle-preserving (i.e., $\forall \x,\y \medspace \inner{\P\x}{\P\y} = \inner{\x}{\y}$), or
		\item $\exists k \geq 0 : \|\P\|_F = k \quad \forall \P \in \mathcal{F}$ and the matrix $\Y$ is $\nu$-whitened (i.e., $\Y\diag{\q}^2\Y^\top = \I_d$),
		\vspace{-.25cm}		
	\end{enumerate}
	Problem \eqref{eq:euclidean_formulation} is equivalent to 
	\begin{equation}\label{eq:simplified_formulation}
		\max_{\Gamma \in \Pi(\p,\q)}\max_{\P \in \cF}  \inner{\X\Gamma\Y^\top}{\P}. 
	\end{equation}
\end{lemma}
Condition (1) above is reasonable as it guarantees $\P$ preserves geometric relations across spaces. On the other hand, whitening is a common pre-processing step in feature learning \citep{hyvarinen2000independent} and correspondence problems \citep{artetxe2018unsupervised}. 

The following generalization of Lemma~\ref{lemma:procrustes_general} shows that when optimizing over Schatten $\ell_p$-norm balls, the inner problem in \eqref{eq:simplified_formulation} admits a closed form solution.

\begin{lemma}\label{lemma:svd_general}
	Let $\M$ be a matrix with SVD  $\M=\U\Sigma\V^\top$ and let $\Sigma = \diag{\pmb{\sigma}}$, then
	\begin{equation}
		\argmax_{\P : \|\P\|_p \leq k} \inner{\P}{\M} = \U\diag{\s}\V^\top
	\end{equation}
	where $\s$ is such that $\|\s\|_p \leq k$ and attains $\s^\top\pmb{\sigma} = k\|\pmb{\sigma}\|_q$, for $\|\cdot\|_q$ the dual norm of $\|\cdot\|_p.$
\end{lemma}
Therefore, the inner problem in \eqref{eq:simplified_formulation} boils down to maximization of support functions of vector-valued $\ell_p$ balls, which can be done in closed form for any $p\geq 1$ by choosing $s_i\propto \sigma_i^{q-1}$ \citep{jaggi2013revisiting}. This, in turn, greatly simplifies the alternating optimization approach.  For a fixed $\Gamma$, we can use Lemma~\ref{lemma:svd_general} to obtain a closed-from solution $\P^*$. On the other hand, for a fixed $\P$, optimizing for $\Gamma$ is a classic discrete optimal transport problem with cost matrix $\tilde{\C} = -\X^\top\P\Y$,\footnote{This is of course equivalent to solving the original problem \eqref{eq:original_formulation}, whose cost matrix has a simpler interpretation.} which can be solved with off-the-shelf OT algorithms.

Next, we investigate what Lemma~\ref{lemma:svd_general} implies for two salient cases, $p=\infty$ and $p=2$. The case $p=1$ is discussed in the Appendix. Then, in  Section~\ref{sub:optim}, we discuss optimization of the general problem in detail.

\subsection{The case $p=\infty$}\label{sub:pinfty}
The Schatten $\ell_\infty$-norm is the spectral norm $\|A\|_{\infty} = \sigma_{\max}(A)$. To guarantee that the identity is contained in $\mathcal{F}_{\infty}$, we take $k_{\infty}\triangleq 1$. Note that combining either condition in Lemma~\ref{lemma:simplifying_conditions} with this implies that $\mathcal{F}_{\infty} = \mathcal{O}_n$.
\iffalse
Indeed, the extreme points of the spectral unit ball are precisely the orthogonal matrices, so optimizing a linear objective such as \eqref{eq:simplified_formulation} over the former or the latter is equivalent.
\fi
Therefore, this choice of norm naturally encodes invariance to rotations and reflections. The dual characterization of Schatten norms implies that
\begin{equation}
	\max_{\P \in \cF_{\infty}}  \inner{\X\Gamma\Y^\top}{\P} = \|\X\Gamma\Y^\top\|_{*}
\end{equation}
so that \eqref{eq:simplified_formulation} becomes a single-block problem:
\begin{equation}\label{eq:nuclear_problem}
	\max_{\Gamma \in \Pi(\p,\q)} \|\X\Gamma\Y^\top\|_*
\end{equation}

Albeit succinct, this alternative representation of the problem is not easier to solve. Despite having eliminated $\P$, the objective is now non-convex with respect to $\Gamma$ (maximization of a convex function). Nevertheless, this formulation offers an interesting geometric interpretation. When $\p,\q$ are uniform distributions, then $\hat{\X}\triangleq \X \Gamma$ is a matrix whose columns correspond to the those of $\X$ transported to $\cY$ according to the optimal barycentric mapping. Hence, $\hat{\X}\Y^{\top}$ is the (shifted) cross-covariance matrix of the features in $\cX$ and $\cY$ space, i.e., $[\hat{\X}\Y^{\top}]_{ij} = \cov(\hat{x}_i, y_j)$, and its norm indicates the strength of correlation of these features. Therefore, problem \eqref{eq:nuclear_problem} essentially seeks a transport coupling that maximizes the correlation of feature dimensions after transportation.
We leave exploration of direct techniques to optimize \eqref{eq:nuclear_problem} for future work. Here instead we rely on the generic alternating minimization scheme described in the previous section. 

\subsection{The case $p=2$}\label{sub:p2}
The Schatten $\ell_2$-norm is the Frobenius norm. Since $\|\I_d\|_{F} = \sqrt{d}$, we take $\cF_{2} = \left \{ \P \suchthat \|\P\|_F = \sqrt{d} \right \}$. As before, we use Schatten norm duality to note that 
\begin{equation*}
	\max_{\P \in \cF_{2}}  \inner{\X\Gamma\Y^\top}{\P} = \sqrt{d}\|\X\Gamma\Y^\top\|_{F},
\end{equation*}
whereupon problem \eqref{eq:simplified_formulation} now becomes
\begin{equation}\label{eq:frobenius_problem}
	\max_{\Gamma \in \Pi(\p,\q)} \|\X\Gamma\Y^\top\|_{F},
\end{equation}
with similar intuition to Problem \eqref{eq:nuclear_problem}, but for a different metric.
However, this subtle difference has important consequences, such as the following connection.

\begin{lemma}\label{lemma:gw}
	Consider the Gromov-Wasserstein problem for discrete measures $\mu$ and $\nu$ \citep{peyre2016gromov}:
\begin{equation}\label{eq:gw_general}
	\min_{\Gamma \in \Pi(\p,\q)} \sum_{i,j,k,l} L(\C_{ik}^x, \C_{jl}^y)\Gamma_{ij}\Gamma_{kl},
\end{equation}
where $(\C^x,\p)$ and $(\C^y,\q)$ are (intra-space) measured similarity matrices and $L$ is a loss function. For the choice of cosine similarity and squared loss $L(a,b) = \frac{1}{2}|a -b|^2$, Problems~\eqref{eq:gw_general} and \eqref{eq:frobenius_problem} are equivalent.
\end{lemma}

\subsection{Optimization}\label{sub:optim}
We solve problem \eqref{eq:simplified_formulation} with alternating maximization on $\Gamma$ and $\P$. For a fixed $\Gamma$,  Lemma~\ref{lemma:svd_general} shows a closed-from solution $\P^*$ at the cost of an $d\times d$ SVD, i.e., $O(d^3)$. For a fixed $\P$, the optimal $\Gamma^*$ can be obtained exactly in $O(N^3\log N)$ time ($N= \max\{m,n\}$) via linear programming, or  $\epsilon$-approximately via the Sinkhorn algorithm in $O(N^2 \log N \epsilon^{-3})$ time \citep{altschuler2017near-linear}. The latter would correspond to solving an entropy-regularized version of the problem:
\begin{equation}\label{eq:entreg_formulation}
\max_{\Gamma \in \Pi(\p,\q)}\max_{\P \in \cF}  \inner{\Gamma}{\X^\top\P\Y}  + \lambda H(\Gamma).     
\end{equation}
The Sinkhorn algorithm can be applied even to the original (non-regularized) problem \eqref{eq:simplified_formulation} by relying on inexact alternating minimization methods that allow for approximate solution of intermediate steps \citep{eckstein2017approximate,mokhtari2015decentralized}. Besides providing an alternative algorithmic approach, this observation could be used to prove convergence rates for problem \eqref{eq:simplified_formulation}. Here, we instead focus on optimizing the regularized formulation \eqref{eq:entreg_formulation}.

Alternating optimization methods for non-convex objectives are known to be sensitive to initialization \citep{jain2013low,hardt2014understanding}. Indeed, a key component of fully unsupervised approaches to feature alignment is finding good quality initializations. For example, for unsupervised word embedding alignment, state-of-the-art methods rely on additional---often heuristic---steps to generate good initial solutions, such as adversarially-trained neural networks \citep{conneau2018word,zhang2017earth,zhang2017adversarial}, which themselves are often very sensitive to initialization, sometimes failing completely on the same problem for different random restarts \citep{artetxe2018unsupervised}.

Neither Problem \eqref{eq:simplified_formulation} nor \eqref{eq:entreg_formulation} is jointly concave in $\Gamma$ and $\P$, thus facing in principle a similar challenge in terms of sensitivity to initialization. However, in \eqref{eq:entreg_formulation} the strength of the entropic regularization controls the extent of non-concavity: strong regularization leads to a more concave objective, while $\lambda \rightarrow 0$ leads to increasingly more non-concave objective. We propose to leverage this observation to alleviate sensitivity to initialization by using an annealing scheme on the regularization term. Starting from a large value of $\lambda$, we decay this value in each iteration by setting $\lambda_t = \alpha \times \lambda_{t-1}$ with $\alpha \in (0,1)$, until a minimum value $\underline{\lambda}$ is reached. We stop the method when the value of the objective converges. The advantage of this annealing approach is that it avoids ad-hoc initialization, and eliminates the need for hyperparameter tuning on $\lambda$, since any sufficiently large choice of $\lambda_0$ achieves the same objective. In \textit{all} our experiments, we use the same parameter values $\lambda_0 = 1$ and decay $\alpha = 0.95$. 

The method described so far, summarized as Algorithm~\ref{alg:invarot} in the supplement, is used in our first set of experiments. However, to scale up to very large problems, such as the motivating application of word embedding alignment, we propose an incremental approach, described in detail in Appendix~\ref{sec:scaling}.

\begin{figure}[t!]
    \centering
    \begin{subfigure}[t]{0.49\linewidth}
        \centering
	    \includegraphics[trim={16cm, 1.8cm, 14.5cm, 3cm}, clip, width=\linewidth]{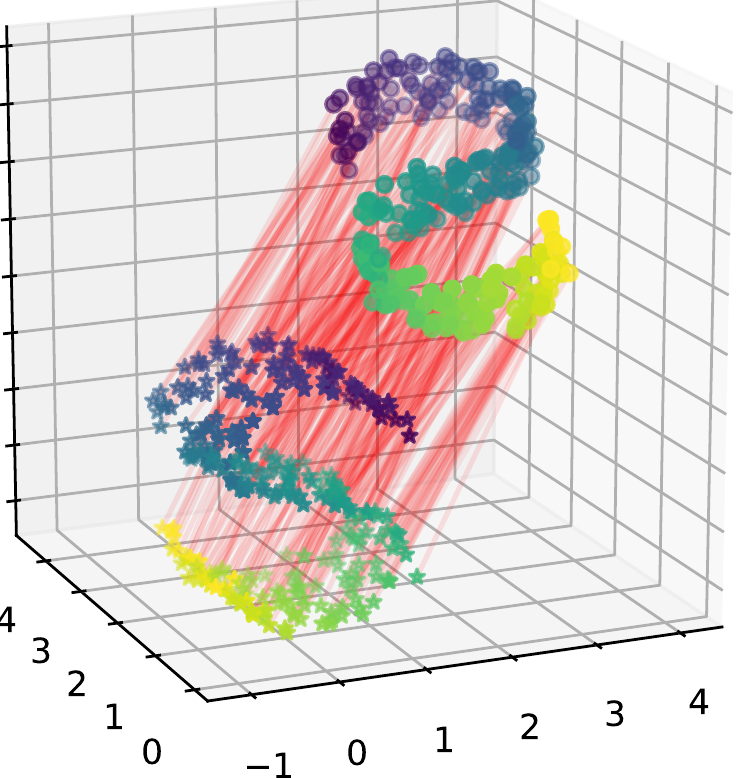}
        \caption{Classic OT}
    \end{subfigure}%
    ~ 
    \begin{subfigure}[t]{0.49\linewidth}
        \centering
	    \includegraphics[trim={25.7cm, 1.8cm, 4.8cm, 3cm}, clip, width=\linewidth]{figs/Sdata3d_Orth}
        \caption{$\ell_{\infty}$-invariant OT}
    \end{subfigure}
    \caption{Optimal couplings for the synthetic point cloud dataset with underlying orthogonal ($\cF_{\infty}$) invariance; green (red) edges denote (in)correct matchings.}\label{fig:synth_matchings}
\end{figure}

\section{Experiments}\label{sec:experiments} 
\vspace{-0.1cm}
\subsection{Synthetic Datasets}

First, we test our approach in a controlled setting with known underlying invariance. We generate a point cloud in $\R^3$ (the \emph{source}), and then apply a transformation $\P$ randomly sampled from one of the families $\cF_p$ to generate a \emph{target} point cloud. The goal is thus to recover the true correspondences between source and target points. We generate a discrete matching $\psi$ from a coupling $\Gamma$ as $\psi(i) = \argmax_j \Gamma_{ij}$, and compute its accuracy with respect to the known true point-wise correspondences. As expected, when the true latent transformation is orthogonal, endowing OT with $\ell_\infty$ invariance allows it to recover the correct matching between the point clouds, while the classic (invariance-agnostic) formulation does not, greedily matching based on proximity instead (Figure~\ref{fig:synth_matchings}). 

\begin{figure}[t!]
	\centering
	\includegraphics[trim={0.4cm, 1cm, 1.5cm, 8.2cm}, clip,width=0.9\linewidth, right]{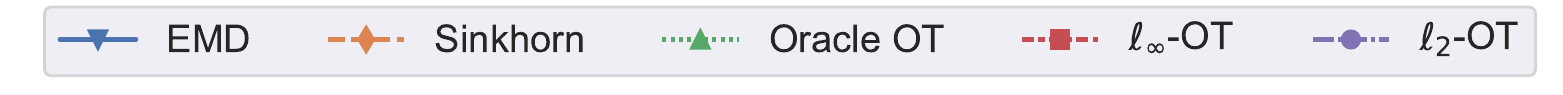}\\
	\begin{subfigure}[t]{0.49\linewidth}
        \centering
        \includegraphics[trim={0.32cm, 3.3cm, 9cm, 1.2cm}, clip,height=3cm]{figs/noise_sensitivity_combined_accuracy_5rep_20start.pdf}\\
    	\includegraphics[trim={0.32cm, 2.3cm, 9cm, 1.07cm}, clip,height=3.55cm]{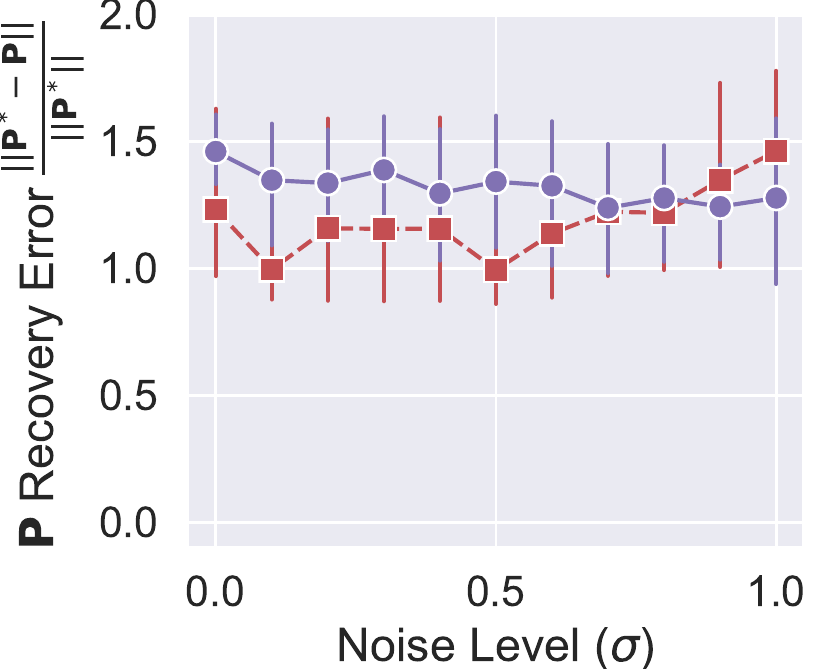}
        \caption{True transform in $\cF_{2}$}
    \end{subfigure}%
    ~ 
	\begin{subfigure}[t]{0.49\linewidth}
        \centering
        \includegraphics[trim={8.75cm, 3.3cm, 1.78cm, 1.2cm}, clip,height=3cm]{figs/noise_sensitivity_combined_accuracy_5rep_20start.pdf}\\
    	\includegraphics[trim={8.75cm, 2.3cm, 1.78cm, 1.07cm}, clip,height=3.55cm]{figs/noise_sensitivity_combined_error_5rep_20start.pdf}
        \caption{True transform in $\cF_{\infty}$}
    \end{subfigure}
	\caption{Results for the synthetic datasets. \textbf{Top}: matching accuracy for the computed coupling $\hat{\Gamma}$. \textbf{Bottom}: error in recovering $\P$. The plots show mean values and one s.d.~error bars over 5 repetitions.}\label{fig:synth_results}
\end{figure}

Next, we investigate the effect of noise. As before, we generate point clouds with two types of invariances ($\cF_{2}$ and $\cF_{\infty}$), but now add a Gaussian noise term with variance $\sigma$ to the target points. We compare the performance of the following versions of OT: classic \textsc{Emd} \eqref{eq:original_OT}, the entropy-regularized formulation \eqref{eq:entropy_reg_ot} solved via the \textsc{Sinkhorn} algorithm, our $\ell_2$ and $\ell_{\infty}$-invariant versions, and an \textsc{Oracle} which solves a entropy-regularized problem \emph{without the transformation applied}, i.e.~only adding noise. Figure~\ref{fig:synth_results} (top) shows the matching accuracy (mean and one standard deviation over 5 repetitions). As expected, $\ell_{\infty}$-OT is better than $\ell_{2}$-OT at recovering the correspondences when the true transformation is orthogonal, and vice versa; and there is a loss of accuracy caused by the estimation of $\P$ in the $\ell_2$ case, as shown by the gap between our methods and \textsc{Oracle OT}. But surprisingly, both invariance methods outperform the oracle in the  $\ell_{\infty}$ case, which we attribute to the added freedom of choosing $\P$ to overcome noise, combined with the ease of optimizing over $\cF_{\infty}$ compared to $\cF_{2}$. This hypothesis is supported by the overall higher error in recovering $\P$ in the latter case (Fig.~\ref{fig:synth_results}, bottom).

\begin{table*}[t]
  \scriptsize
  \centering
  \ra{1.3}
  \resizebox{\linewidth}{!}{%
  \begin{tabular}{r l c c c c c c c c c c c c c c}\toprule
 	   & & & \multicolumn{2}{c}{\En-\Es} & \multicolumn{2}{c}{\En-\Fr}  & \multicolumn{2}{c}{\En-\De}  & \multicolumn{2}{c}{\En-\It}  & \multicolumn{2}{c}{\En-\Ru} \\
	   \cmidrule(lr){4-5} \cmidrule(lr){6-7}  \cmidrule(lr){8-9}  \cmidrule(lr){10-11}  \cmidrule(lr){12-13}
	   & Supervision & Time & $\rightarrow$ & $\leftarrow$ & $\rightarrow$ & $\leftarrow$ & $\rightarrow$ & $\leftarrow$ & $\rightarrow$ & $\leftarrow$ & $\rightarrow$ & $\leftarrow$  \\
	    \midrule
		\procru  & 5K words & 3 & 77.6 & 77.2 & 74.9 & 75.9 & 68.4 & 67.7 & 73.9 & 73.8 & 47.2 & 58.2\\
		\procru + \textsc{Csls} & 5K words & 3 & 81.2 & 82.3 & 81.2 & 82.2 & 73.6 & 71.9 & 76.3 & \textbf{75.5} & 51.7 & 63.7 \\
        \textsc{Adv + Csls} &  None  & 643 & 75.7 & 79.7 & 77.8 & 71.2 & 70.1 & 66.4 & 72.4 & 71.2 & 37.1 & 48.1 \\
	    \textsc{Adv + Csls + Refine} &  None  & 957 & 81.7 & 83.3 & 82.3 & 82.1 & 74.0 & 72.2 & 77.4 & 76.1& 44.0 & 59.1 \\
	    \textsc{Wasserstein + Csls} &  None  & -- & 82.8 & 84.1 & 82.6 & 82.9 & 75.4 & 73.3 & -- & -- & 43.7 & 59.1 \\
		\textsc{Gromov-Wasserstein} & None & 37 & 81.7 & 80.4 & 81.3 & 78.9 & 71.9 & 72.8 & 78.9 & 75.2 & 45.1 & 43.7  \\
		\textsc{Self-Learn + Csls} & None & 476 & 82.3 & 84.7 & 82.3 & 83.6 & 75.1 & 74.3 & 79.2 & 79.8 & 48.9 & 65.9  \\
		\midrule		
		$\ell_{\infty}$-\textsc{InvarOt + Csls}&  None & 70 & 81.3 & 81.8 & 82.9 & 81.6 & 73.8 & 71.1 & 77.7  & 77.7 & 41.7 & 55.4 \\		
    \bottomrule
  \end{tabular}%
}
  \caption{Accuracy on the word translation task.  ``Time'' refers to average runtime in minutes per problem (language pair) on the same \textbf{CPU} machine. We report results for our $\ell_{\infty}$-OT method \emph{without} relying on the iterative refinement step of \citet{conneau2018word}, so it is more appropriately compared to their \textsc{Adv + Csls}.}
  \label{tab:conneau_results}
\end{table*}

\subsection{Unsupervised word translation} %
\label{sub:unsup_word_translation}

\vspace{-0.2cm}

Automatic word translation has a long tradition (under the name \emph{bilingual lexical induction}) in computational linguistics \citep{rapp1995identifying, fung1995compiling}, and has seen a recent revival due to the success of fully unsupervised methods \citep{conneau2018word, Artetxe2018Robust}. Most such methods cast the problem as feature alignment between sets of word embeddings, motivated by the observation that these possess similar geometry across languages \citep{Mikolov2013Exploiting}. Though their relational structure is similar, the absolute position of these vectors is irrelevant. Indeed, word embedding algorithms are naturally interpreted as metric recovery methods \citep{Hashimoto2016Word}, making these vectors intrinsically invariant to angle (or distance) preserving transformations. This observation suggests inducing invariance to orthogonal transformations, as described in Section~\ref{sub:pinfty}.

Most current unsupervised methods circumvent this issue by resorting to ad-hoc normalization, joint re-embedding, or by estimating a complex mapping between the two spaces with adversarial training. These methods require careful initialization and post-mapping refinements, such as mitigating the effect of frequent words on neighborhoods, and are often hard to tune properly \citep{Artetxe2018Robust}. 

\subsubsection{Dataset and Methods}
\vspace{-0.15cm}
The dataset of \citet{conneau2018word} consists of fastText word embeddings and parallel dictionaries for 110 language pairs. Here, we focus on the five language pairs for which they report results: English to Spanish, French, Italian, German and Russian. As they do, we include a strong semi-supervised baseline of solving a \procru problem directly using the available (5K) cross-lingual embedding pairs. In addition, we compare against various unsupervised methods: \textsc{SelfLearn} \citep{Artetxe2018Robust}, \textsc{Gromov-Wasserstein} \citep{alvarez-melis2018gromov}, \textsc{Adv} \citep{conneau2018word} and \textsc{Wasserstein} \citep{grave2018unsupervised}.\footnote{Despite its relevance, we do not include the OT-based method of \citet{zhang2017earth} in the comparison because their implementation required use of proprietary software.} The code for the last of these was not available, so we report results from their paper (which excludes \textsc{En}$-$\textsc{It}), and omit runtime. Whenever nearest neighbor search is required, we use Cross-domain Similarity Local Scaling (\textsc{Csls}) \citep{conneau2018word}.

\subsubsection{Results} %
\begin{figure}[t]
	\centering
	\includegraphics[scale=0.28, trim = {0.2cm, 0.35cm, 20cm, 0.35cm}, clip]{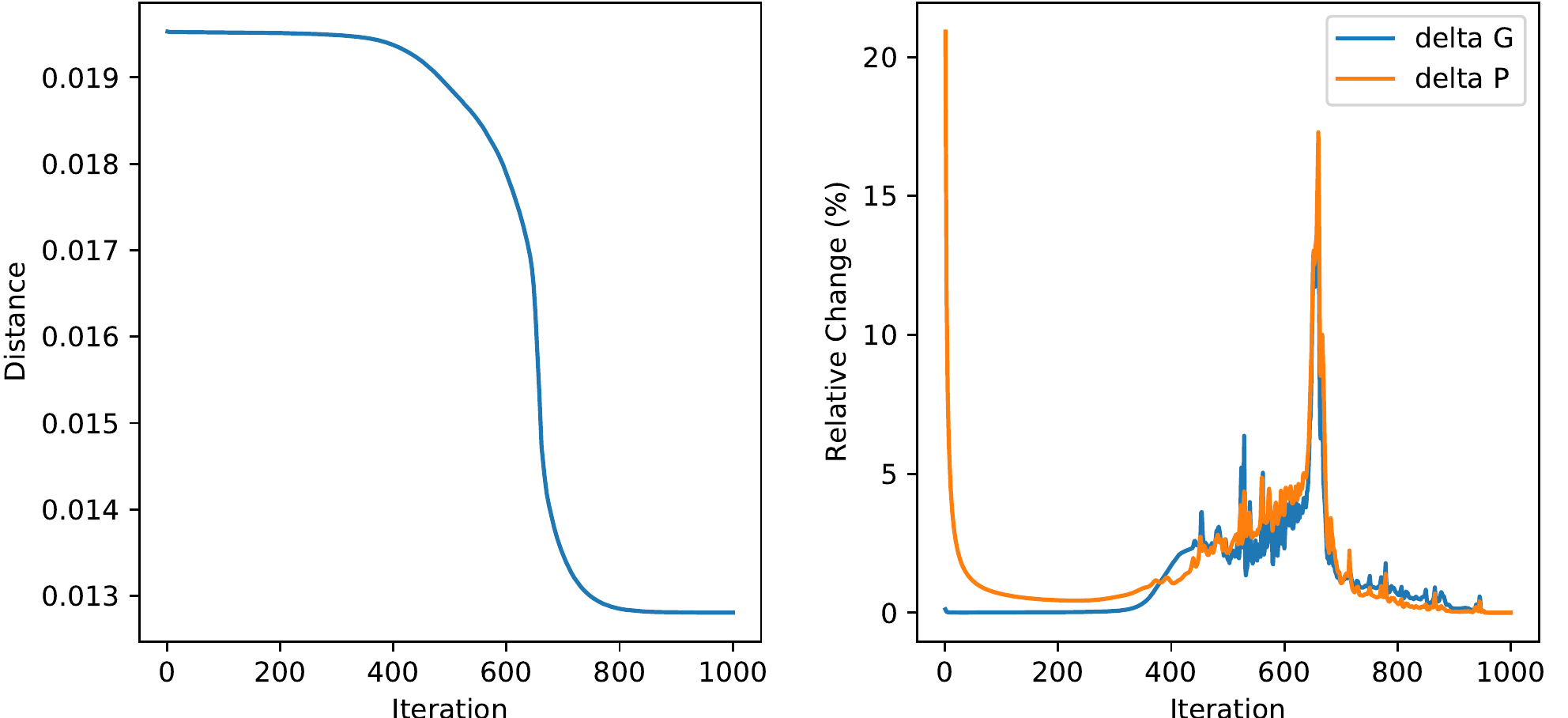}%
	\includegraphics[scale=0.28, trim = {30.5cm, 0.35cm, 0cm, 0.35}, clip]{figs/procOT_enit_5000_history.pdf}	
	\caption{Optimization dynamics of $\ell_{\infty}$-OT on the word translation task (\textsc{En}$\rightarrow$\It, 5K vocab). Left to right: objective, change in $\P$ and $\Gamma$ from previous iteration, and translation accuracy (Precision@K).}\label{fig:dynamics}
\end{figure}

\label{sec:results}
\paragraph{Optimization Dynamics.} 
Figure~\ref{fig:dynamics} shows a typical run of our algorithm for $\ell_{\infty}$-OT, exhibiting a common pattern: little progress at the beginning (during which $\P$ is being aggressively adjusted), followed by a steep decline in the objective (during which both $\P$ and $\Gamma$ are increasingly modified in each step), after which convergence is reached. Note how the value of the optimization objective (left) and the accuracy in the translation task (right) are strongly correlated, particularly when compared against adversarial networks \citep{conneau2018word}. This is crucial because accuracy (shown here for expository purposes) \textit{is not available} during the actual task, so model selection and early stopping are made based solely on the unsupervised objective. In addition, note that except for a small adjustment at the end of training, our method does not risk degradation by over-training, as is often the case for adversarial training alternatives.

\vspace{-0.2cm}
\paragraph{Performance.} The accuracy results (Table~\ref{tab:conneau_results}) show that our general framework performs on par with state-of-the-art approaches tailored this task, at a fraction of the computational cost, and yields an objective which is more faithful to the true metric of interest.

\section{Discussion}
\vspace{-0.2cm}
We proposed a general formulation of optimal transport that accounts for global invariances in the underlying feature spaces, unifying various existing approaches to deal with such invariances. The problem allows for very efficient algorithms in two cases often found in practice. The experiments show that this framework provides a fast, principled and robust alternative to state-of-the-art methods for unsupervised word translation, delivering comparable performance. These results suggest that OT with invariances is a viable alternative to adversarial methods that infer correspondences from complex, often underdetermined, neural network maps.

\clearpage
\pagebreak

\section*{Acknowledgements}
The authors would like to thank Youssef Mroueh and Suvrit Sra for helpful discussions and remarks. The work was partially supported by the MIT-IBM project on adversarial learning of multi-modal and structured data, and by NSF CAREER award 1553284.

\printbibliography

\clearpage
\pagebreak

\appendix

\section{Dealing with different samples sizes and dimensions}

In the general case where we consider collections of potentially different size ($n\neq m$) of vectors of potentially different dimensionality ($d_x \neq d_y$), we have that $\X$ and $\Y$ are matrices of size $d_x \times m$ and $d_y \times n$, respectively, while $\P$ is of size $d_x \times d_y$. 

Note that for the transportation problem in \eqref{eq:simplified_formulation}, i.e.,
$$\max_{\Gamma \in \Pi(\p,\q)}  \inner{\Gamma}{\X^{\top}\P\Y},$$
the dimensions $d_x, d_y$ are irrelevant. While the transportation coupling $\pmb{\Gamma}$ is now of size $m \times n$, the problem is equally meaningful as before, i.e., DOT is well formulated and solved analogously in this case ($n\neq m)$ as in the case of equal-sized marginals.

On the other hand, the transformation problem in \eqref{eq:simplified_formulation}, namely
$$\max_{\P \in \cF}  \inner{\X \Gamma \Y^{\top}}{\P},$$
is oblivious to $n$ and $m$. However, when $d_x \neq d_y$, the problem no longer admits a closed form solution in general, so in this case this step requires optimization too. However, there exist various iterative algorithms to solve this problem---known as \textit{unbalanced Procrustes}---efficiently \citep{gower2004procrustes, park1991parallel, viklands2006algorithms}. 

\section{The case $p=1$}\label{sub:p_1}

Recall that the Schatten $\ell_1$-norm is the nuclear norm $\|A\|_* = \sum_{i=1}^n \sigma_i(A)$. Therefore, the invariance set of interest is now
\begin{equation}
	\cF_{1} = \left \{ \P \suchthat \|\P\|_{*} = d \right \},
\end{equation}
which, as before, contains the identity matrix.
Note that adding either condition in Lemma~\ref{lemma:simplifying_conditions} yields, again, the set of orthonormal matrices.\footnote{The intersection of the Schatten $\ell_2$ and $\ell_\infty$ norm balls, defined in terms of that of the $\ell_2$ and $\ell_{\infty}$ vector norm balls, occurs in the extreme points of the latter (see Fig.~\ref{fig:schatten}).} Therefore, in the case one wants to rely on Lemma~\ref{lemma:svd_general} to solve the problem efficiently, this choice of invariance ends up being equivalent to the $p=\infty$ case described in Section~\ref{sub:pinfty}. However, we remark that this equivalence is a consequence of the simplifying assumptions, and that one could still solve this problem with the Frank-Wolfe approach described in Section~\ref{sub:optim}, in which case the two cases $p=\infty$ and $p=1$ would indeed lead to different solutions.

\section{Further Extensions}

The framework proposed here can be further extended by considering other transformations that can be easily incorporated into the Procrustes problem framework. For example, scaling and translation can be added on top of orthogonal Procrustes and still yield a closed form solution \citep{gower2004procrustes}.

\section{Proofs}

\begin{customlemma}{\ref{lemma:simplifying_conditions}}
	If any of the following conditions holds;
	\begin{enumerate}
		\item $\forall \P \in \mathcal{F}$, $\P$ is angle-preserving
		\item $\exists k \geq 0 : \|\P\|_F = k \quad \forall \P \in \mathcal{F}$ and the matrix $\Y$ is $\nu$-whitened (i.e., $\Y\diag{\q}^2\Y' = \I_d$).
	\end{enumerate}
	then problem \eqref{eq:euclidean_formulation} is equivalent to \begin{equation}\label{eq:simplified_formulation_appendix}
		\max_{\Gamma \in \Pi(\mu,\q)}\max_{\P \in \cF}  \inner{\Gamma}{\X'\P\Y} = 		\max_{\Gamma \in \Pi(\p,\q)}\max_{\P \in \cF}  \inner{\X\Gamma\Y'}{\P} 
	\end{equation}
    \begin{proof}
    	Suppose (1) holds, i.e., $\inner{\P\x}{\P\y} = \inner{\x}{\y}$ for every $\x,\y\in \R^d$. Then, in particular $\|\P\y\|_2 = \|\y\|_2$ for every $\y^{(j)}$, and therefore:
        \[ \inner{\v}{\q} = \sum_{j=1}^{m} \|\P\y^{(j)}\|_2 = \|\y^{(j)}\|_2\]
        and therefore only the first term in \eqref{eq:simplified_formulation} depends on $\P$ or $\Gamma$, from which the conclusion follows. On the other hand, suppose (2) holds, and let $\tilde{\Y}=\Y\diag{\q}$, so that $\tilde{\Y}\tilde{\Y}'=\I_d$. We have:
       \begin{align*}
       \inner{\v}{\q} &= \sum_{i=1}^m q_j \|\P \y^{(j)}\|_2^2 \\ &= \sum_{j=1}^m \|\P\y^{(j)}q_j\|_2^2 \\&= \|\P\tilde{\Y}\|_2^2 \\&= \inner{\P\tilde{\Y}}{\P\tilde{\Y}} = \inner{\P}{\P\tilde{\Y}\tilde{\Y}'} =  \|\P\|_F^2 = k^2,
       \end{align*}
       that is, $\inner{\v}{\q}$ again does not depend on $\P$. This concludes the proof.
    \end{proof}
\end{customlemma}

\begin{customlemma}{\ref{lemma:svd_general}}
	Let $\M$ be a matrix with SVD decomposition $\M=\U\Sigma\V'$ and let $\Sigma = \diag{\pmb{\sigma}}$, then
	\begin{equation}\label{eq:svd_general_appendix}
		\argmax_{\P : \|\P\|_p \leq k} \inner{\P}{\M} = \U\diag{\s}\V'
	\end{equation}
	where $\s$ is such that $\|\s\|_p \leq k$ and attains $\s'\pmb{\sigma} = k\|\pmb{\sigma}\|_q$, for $\|\cdot\|_q$ the dual norm of $\|\cdot\|_p$. 
    \begin{proof}
     Suppose $\P$ is such that $\|\P\|_p\leq k$, and let $\U_\P\diag{\s}\V_{\P}'$ be its singular value decomposition. This implies that $\|\s\|_p = \|\P\| \leq k$.  In addition,
    	\begin{align*}
    		\inner{\P}{\M} &=\inner{\P}{\U\Sigma\V'} \\ &=  \inner{\U'\P\V}{\Sigma} \\ &= \sum_{i=1}^d [\U'\P\V]_{ii} \sigma_i(\M) \\ &= \sum_{i=1}^d \u_i\P\v_i\sigma_i(\M) \leq \sum_{i=1}^d s_i\sigma_i(\M) = \inner{\s}{\pmb{\sigma}}
    	\end{align*}
        Here, the inequality holds because, by definition of the SVD decomposition, for every $i$ it must hold that $\|\u_i\|_2 = \|\v_i\|_2=1$ and
\begin{equation}\label{eq:svd_variational_bound}
     \u_i\P\v_i \leq \sup_{\substack{\u \perp \text{span}\{\u_1,\dots,\u_{i-1}\} \\  \v \perp \text{span}\{\v_1,\dots,\v_{i-1}\} }} \frac{\u'\P\v}{\|\u\|\|\v\|} \leq \sigma_i(\P) = s_i
\end{equation}
      Therefore:
      \begin{align*}
        \sup_{\P : \|\P\|_p \leq k} \inner{\P}{\M} & \leq \sup_{\s : \|\s\|_p \leq k } \inner{\s}{\pmb{\sigma}}  \\
        &= k\sup_{\s : \|\s\|_p \leq 1 } \inner{\s}{\pmb{\sigma}} = k\|\pmb{\sigma} \|_q           
      \end{align*}
        where the last equality follows from the definition of dual norm for vectors.
        
        Conversely, take any vector $\s$ with $\|\s\|_p = k$, and define $\tilde{\P}(\s) = \U \diag{\s}\V'$. Clearly, $\|\tilde{\P}(\s)\|_p = k$, so the supremum must satisfy:
        \begin{align*}
        	\sup_{\P} \inner{\P}{\M} &\geq \sup_{\s : \|\s\|_p\leq k}\inner{\tilde{\P}(\s)}{\M} \\&= \sup_{\s : \|\s\|_p\leq k}\inner{\U\diag{\s}\V'}{\U\Sigma \V'} \\&= \sup_{\s : \|\s\|_p\leq k} \inner{\diag{\s}}{\Sigma} = k \|\pmb{\sigma}\|_q
        \end{align*}
        Therefore, we conclude that the optimal value of \eqref{eq:svd_general_appendix} is exactly $k\|\pmb{\sigma}\|_q$. 
        
        Furthermore, \eqref{eq:svd_variational_bound} holds with equality if and only if $(\u_i,\v_i)$ coincide with the left and right singular vectors of $\P$. Thus, any $\P$ maximizing \eqref{eq:svd_general_appendix} must have the form $\P = \U\diag{\s}\V'$, with $\|\s\|_p \leq k$ and $\inner{\s}{\pmb{\sigma}}=k\|\pmb{\sigma}\|_q$, as stated.      
    \end{proof}
\end{customlemma}

\begin{customlemma}{\ref{lemma:gw}}
	Consider the Gromov-Wasserstein problem for discrete measures $\mu$ and $\nu$ \citep{peyre2016gromov}:
\begin{equation}\label{eq:gw_general_app}
	\min_{\Gamma \in \Pi(\p,\q)} \sum_{i,j,k,l} L(\C_{ik}^x, \C_{jl}^y)\Gamma_{ij}\Gamma_{kl},
\end{equation}
where $(\C^x,\p)$ and $(\C^y,\q)$ are (intra-space) measured similarity matrices and $L$ is a loss function. For the choice of cosine similarity and squared loss $L(a,b) = \frac{1}{2}|a -b|^2$, Problems~\eqref{eq:gw_general} and \eqref{eq:frobenius_problem} are equivalent.

\begin{proof}
For the choice of cosine metric, and assuming without loss of generality that the columns of $\X$ and $\Y$ are normalized, the similarity matrices are given by $\C^x = \X^\top\X$ and $\C^y = \Y^\top\Y$. In addition, let $L$ be the $\ell_2$ loss, i.e., $L(a,b) = |a-b|^2$. Then the objective in problem \eqref{eq:gw_general_app} becomes:
\begin{align*}
	\mathcal{L}(\Gamma) &= \sum_{i,j,k,l}\bigl(\C_{ik}^x - \C_{jl}^y \bigr)^2 \Gamma_{ij}\Gamma_{kl} \\ &= \sum_{i,j,k,l}\bigl(\C_{ik}^x)^2 \Gamma_{ij}\Gamma_{kl} - 2\sum_{i,j,k,l}\bigl(\C_{ik}^x\C_{jl}^y \bigr) \Gamma_{ij}\Gamma_{kl} \\ &+ \sum_{i,j,k,l}\bigl(\C_{jl}^y \bigr)^2 \Gamma_{ij}\Gamma_{kl} 
\end{align*}
Since $\Gamma \in \Pi(\p, \q)$, the first of these terms becomes
\begin{equation*}
	\sum_{i,k} \bigl(\C_{ik}^x)^2 \sum_{j,l}\Gamma_{ij}\Gamma_{jl} = \sum_{i,k} \bigl(\C_{ik}^x)^2 \p_i \p_k = \p^\top(\C^x)^2\p
\end{equation*}
where $\p$ is the vector of probabilities in empirical distribution $\mu$, and the last equation follows from the definition of the transportation polytope. Crucially, this term does not depend on $\Gamma$ anymore. Analogously, the last term in $\mathcal{L}(\Gamma)$ does not depend on $\Gamma$ either, so 
\begin{equation}\label{eq:gw_simplified}
	\argmin_{\Gamma \in \Pi(\p,\q)} \mathcal{L}(\Gamma) = \argmax_{\Gamma \in \Pi(\p,\q)}  \sum_{i,j,k,l}\bigl(\C_{ik}^x\C_{jl}^y \bigr) \Gamma_{ij}\Gamma_{kl}
\end{equation}
On the other hand, consider problem \eqref{eq:frobenius_problem}. The objective it seeks to maximize is
\begin{align*}
\|\X\Gamma\Y^\top\|_{F}^2 &= \inner{\X\Gamma\Y^\top}{\X\Gamma\Y^\top} \\ &= \inner{\X^\top\X\Gamma}{\Gamma\Y\Y^\top}\\
&= \sum_{i=1}^n\sum_{l=1}^m \bigl[\X^\top\X\Gamma \bigr]_{il}\bigl[\Gamma\Y^\top\Y\bigr]_{il} \\
&= \sum_{i=1}^n\sum_{l=1}^m \bigl[\C^x\Gamma \bigr]_{il}\bigl[\Gamma\C^y\bigr]_{il} \\ 
&= \sum_{i=1}^n\sum_{l=1}^m \left(\sum_{k=1}^n \C^x_{ik}\Gamma_{kl} \right) \left(\sum_{j=1}^m \Gamma_{ij}\C^y_{jl} \right) \\
&= \sum_{i=1}^n\sum_{l=1}^m \sum_{k=1}^n \sum_{j=1}^m \C^x_{ik}\Gamma_{kl} \Gamma_{ij}\C^y_{jl}
   \end{align*}
   which is exactly the objective in \eqref{eq:gw_simplified}. Hence, Problems \eqref{eq:frobenius_problem} and \eqref{eq:gw_general_app} are indeed equivalent.

    \end{proof}
\end{customlemma}

\section{The Algorithm}\label{sec:algorithm}

\begin{algorithm}[H]
    \caption{Optimal Transport with Invariances}\label{alg:invarot}
\begin{algorithmic}
    \STATE {\bfseries Inputs:}
    \begin{itemize}
	    \vspace{-.3cm}
	    \itemsep-0.3em    
	    
        \item Data matrices and histograms $(\X,\p)$, $(\Y,\q)$
        \item Order of invariance $p$ and radius $k_p$
        \item Initial/final entropy regularization $\lambda_0$ and $\underline{\lambda}$, decay rate $\eta$
    \end{itemize}
   \STATE // Initialize feasible transformation in $\cF_p$
	\STATE $\U, \pmb{\Sigma}, \V^\top \gets \text{SVD}(\textsc{RandomMatrix}(d\times d))$
	\STATE $\pmb{\sigma} \gets \diag{\pmb{\Sigma}}$
	\STATE $\s \gets k_p \cdot \pmb{\sigma}/\|\pmb{\sigma}\|_p$
    \STATE $\P = \U\diag{\s}\V^\top$
   \STATE $\lambda \gets \lambda_0$
	   \WHILE{not converged}
	   \STATE // Compute distances w.r.t.~current mapping $\P$
	   \STATE $\C_{\P} \gets \textsc{PairwiseDistances}(\X,\P\Y)$
	    \STATE // Solve regularized OT via Sinkhorn iterations
		\STATE $\b \gets \ones, \quad \K \gets \exp\{-\C_{\P} /\lambda\}$
	    \WHILE{not converged}
			\STATE $\a \gets \p \oslash \K \b$
			\STATE $\b \gets \q \oslash \K^\top \a$
		\ENDWHILE
	 	\STATE $\pmb{\Gamma} \gets \diag{\a}\K\diag{\b}$
	 	\STATE // Solve generalized Procrustes problem
		\STATE $\U, \pmb{\Sigma}, \V^\top \gets \text{SVD}(\X \pmb{\Gamma}\Y^\top)$
		\STATE $\pmb{\sigma} \gets \diag{\pmb{\Sigma}}$
		\STATE $q \gets \frac{p}{p-1}$
		\STATE $\s \gets k_p \cdot \pmb{\sigma}^{q-1}/\|\pmb{\sigma}^{q-1}\|_p$
    	\STATE $\P = \U\diag{\s}\V^\top$
	   \STATE // Anneal entropy regularization
	   \STATE $\lambda \gets \max\{\lambda * \eta, \underline{\lambda}\}$
	    \ENDWHILE
	\RETURN $\pmb{\Gamma}, \P$
\end{algorithmic}
\end{algorithm}

\section{Solving very large problems}\label{sec:scaling}

 While direct application of Algorithm~\ref{alg:invarot} leads to high-quality solutions for small and mid-sized problems, scaling up to very large sets of points---e.g., hundreds of thousands of word embeddings in the word translation application---can be prohibitive. 
 
 We address this issue by dividing the problem into two phases. In the first stage, we solve a smaller problem (by taking a subsample of $k$ points on each domain thus leading to smaller $\Gamma$ and faster OT solution, but same size of $\P$). Once the first phase reaches convergence, we use the solution $\P^*$ of the first stage to initialize the full-size problem. Note that while this might resemble other approaches that also consider a reduced set of points in their initialization step \citep{conneau2018word, grave2018unsupervised}, a crucial difference is that here we rely on the same optimization problem \eqref{eq:entreg_formulation} in both stages, although with different problem sizes. 
 
 We experimented with various choices of parameter $k$, and observed that the algorithm is remarkably robust to the choice of this parameter. We conjecture that the ordering in which word embeddings are provided (higher-frequency words first, in every language) helps ensure that the solution of the initial problem of reduced size is consistent with the full-size problem.\footnote{This, in fact, points to an issue mostly ignored in previous work on this task: the order of the word embeddings \textit{leaks} important---albeit noisy---correspondence information, which various methods presented as 'fully-unsupervised' seem to rely on one way or another, yet rarely acknowledge it.}
 
 While the end performance is consistent regardless of the choice of sub-sample size $k$, there is naturally a trade-off in run time of the two stages. While solving a smaller initial problem is obviously faster, we observed that in such cases the second stage required more iterations to converge, suggesting that the initial $\P^*$ fed into the second stage was of lower quality (further from the optimal for the full-size problem). In the results presented in Section~\ref{sub:unsup_word_translation}, we take $k$ as large as possible while keeping the time-per-iteration reasonable: $k=5000$. 
 
Note that this strategy of \textit{bootstrapping} solutions of smaller problems can be applied repeatedly, to increasingly grow the problem size over multiple stages. While we did not require to do so in our experiments, it might be an appealing approach for solving extremely large problems.

\end{document}